\documentclass{article}

\usepackage[eandd,preprint]{neurips_2026}

\usepackage[utf8]{inputenc}
\usepackage[T1]{fontenc}
\usepackage{hyperref}
\usepackage{url}
\usepackage{booktabs}
\usepackage{amsfonts}
\usepackage{amsmath}
\usepackage{amssymb}
\usepackage{nicefrac}
\usepackage{microtype}
\usepackage{xcolor}
\usepackage{graphicx}

\title{Active Causal Discovery Benchmark:\\
Evaluating LLM Agents Under Budgeted Interventions}

\author{%
  Sagar Deb \\
  QpiAI \\
  \texttt{sagar.deb@qpiai.tech}
  \And
  Devam Shah \\
  QpiAI \\
  \texttt{devam.s@qpiai.tech}
  \And
  Ashwanth Krishnan \\
  QpiAI \\
  \texttt{ashwanth.krishnan@qpiai.tech}
}

\begin{document}
\maketitle

\begin{abstract}
We introduce the Active Causal Discovery Benchmark (ACDB), an SCM-grounded environment for evaluating whether LLM agents recover causal graph structure from observations and budget-constrained hard interventions. ACDB pairs a linear--Gaussian world generator with a fixed observe--intervene--submit API and a three-layer scoring contract that separates skeleton recovery, DAG recovery, and intervention efficiency. On the current six-level ladder, PC with a greedy active orientation heuristic is the strongest non-oracle method (directed F1 $42.7\%$, SHD $4.79$), ahead of Claude Sonnet 4.6 raw active ($31.7\%$, $7.25$) and GPT-5.4 raw active ($22.9\%$, $9.27$). The most informative diagnostic is the precision--recall decomposition: PC under-commits with high precision, LLMs over-commit with lower precision, and statistical-tool access often increases abstention rather than useful intervention. A structure-blind random DAG baseline reaches $23.6\%$ directed F1 on this dense v0 ladder; a density probe lowers this floor to $16.9\%$, motivating the v1 calibration pass. The current results should therefore be read as a benchmark audit and calibration report, not as evidence that current LLMs solve active causal discovery.
\end{abstract}

\section{Introduction}
\label{sec:introduction}

Causal discovery --- recovering the directed acyclic graph (DAG) behind an observed joint distribution --- is a prerequisite for many forms of causal reasoning. Classical methods such as PC and GES have clean guarantees under faithfulness and causal sufficiency, but they are brittle at small samples and do not by themselves specify how an agent should allocate a finite intervention budget.

Large language models (LLMs) are a natural stress case for this active setting. They can in principle inspect data, choose tests, intervene, and decide when to submit; it remains unclear whether this interaction produces reliable structure recovery or plausible-looking graphs driven by surface regularities.

We ask:
\begin{quote}
\emph{Can an LLM agent, given a budget of hard single-node interventions, recover a causal DAG better than a structure-blind random baseline --- and if so, by how much, and under what conditions?}
\end{quote}

To answer it we introduce \textbf{ACDB}, an Active Causal Discovery Benchmark. ACDB specifies a linear--Gaussian SCM world generator, a push-based session API, a three-tier scoring contract, and a six-level difficulty ladder. Every agent --- classical or LLM --- sees the same paired instances and is scored by the same contract.

The current results support three findings. First, PC with a greedy active orientation heuristic is the strongest non-oracle method: directed F1 $42.7\%$, SHD $4.79$. Second, precision and recall expose different failure modes: PC under-commits with high precision ($66.9\%$) and low recall ($32.8\%$), whereas LLMs over-commit with lower precision and comparable recall. Third, a structure-blind random DAG baseline reaches $23.6\%$ directed F1 on the dense current ladder, so weak LLM margins must be read against a high random floor.

Two secondary findings refine the picture. Statistical-primitive tools (correlation, partial correlation, Fisher-$z$ tests) did not improve directed F1; they often shifted the model toward more testing and less intervention, producing high efficiency through abstention. At L2 and L5, Sonnet 4.6's observational directed F1 exceeded its active-panel directed F1 on the same seeds, consistent with useful observational orientations being revised after intervention data.

The contribution is not a new causal discovery algorithm. It is a benchmark environment and a controlled empirical audit of how current LLM agents behave inside it. ACDB is intentionally narrow: causal sufficiency, linear--Gaussian SCMs, and perfect hard interventions. These restrictions make the scoring contract identifiable and give classical methods a fair comparison point. The density-leakage probe and full correlation-summary ladder define the next calibration work.

\section{Background}
\label{sec:background}

We review only the formal machinery needed to define the benchmark and interpret its scores. Readers familiar with causal discovery may skip this section.

\paragraph{Structural causal models.} A linear--Gaussian structural causal model (SCM) over $X_1, \dots, X_d$ is
\begin{equation}
X_i \;=\; \sum_{j \in \mathrm{Pa}(i)} w_{ij} X_j + \varepsilon_i,
\qquad \varepsilon_i \sim \mathcal{N}(0, \sigma^2_i),
\label{eq:linear-gaussian}
\end{equation}
with $\mathrm{Pa}(i)$ the parents of $X_i$ in a DAG $G$ and independent noise. The SCM induces an observational distribution $P(X)$ and, for every target $(X_i, v)$, an interventional distribution $P(X_{\neq i} \mid \mathrm{do}(X_i = v))$ obtained by replacing $X_i$'s structural equation with $X_i := v$.

\paragraph{Markov equivalence and the CPDAG ceiling.} Two DAGs are \emph{Markov equivalent} iff they share the same skeleton and the same v-structures~\citep{verma1990equivalence,andersson1997characterization}. The equivalence class is represented by a completed partially directed acyclic graph (CPDAG), which orients all edges whose direction is identifiable from $P(X)$ (\emph{compelled} edges) and leaves the rest undirected. An agent working from observations alone is bounded by the CPDAG; to recover the true DAG within an equivalence class, it must rely on structural assumptions or on interventions.

\begin{figure}[t]
\centering
\includegraphics[width=\linewidth]{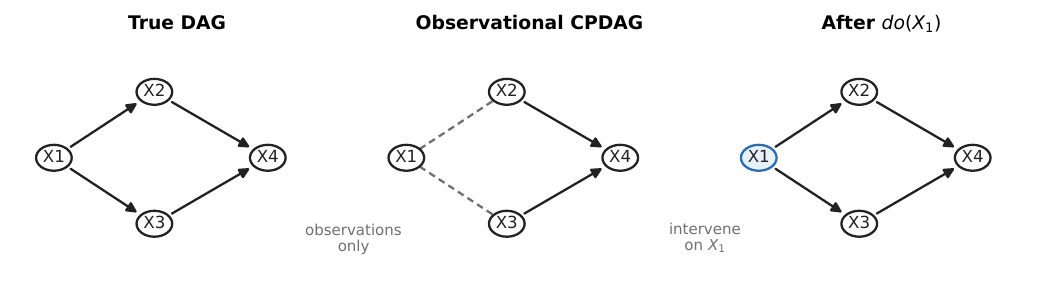}
\caption{Observation identifies a CPDAG, not necessarily the full DAG. In this example the collider into $X_4$ is compelled from observational structure, while the two incident edges at $X_1$ remain undirected until an intervention on $X_1$ resolves them.}
\label{fig:dag-cpdag-intervention}
\vspace{-0.6em}
\end{figure}

\paragraph{Faithfulness and causal sufficiency.} \emph{Faithfulness} states $X_i \perp\!\!\!\perp X_j \mid S \iff X_i$ is $d$-separated from $X_j$ given $S$ in $G$; violations form a measure-zero subset of the parameter space but are empirically troublesome at finite samples, so our world generator rejects instances that approach the boundary. \emph{Causal sufficiency} says that every common cause of observed variables is itself observed. v0 assumes both, plus linear--Gaussian noise. These restrictions rule out latent confounders and nonlinear mechanisms; they are deliberate because they make the scoring contract identifiable and PC a meaningful baseline.

\paragraph{The PC algorithm.} PC~\citep{spirtes2000causation} tests conditional independences $X_i \perp\!\!\!\perp X_j \mid S$ for increasing $|S|$, removes adjacencies, and orients edges via v-structure detection and Meek's rules. Under causal sufficiency, faithfulness, and a correct independence oracle, it returns the true CPDAG; with finite samples it returns an empirical approximation dependent on the chosen test and significance level.

An active causal discovery agent therefore has two distinct jobs: (i) recover the skeleton and compelled orientations, a task bounded by $P(X)$; and (ii) resolve the remaining undirected edges using interventions, a task that scales with the quality of the intervention policy. Our scoring contract (\S\ref{sec:benchmark}) measures both separately.

\section{The ACDB Benchmark}
\label{sec:benchmark}

\textbf{ACDB} is an interactive benchmark environment in which an agent receives observational data, spends a budget of hard single-node interventions, and submits an estimated causal graph. A benchmark instance is $(G, \mathcal{M}, B, D^{\mathrm{obs}})$: a ground-truth DAG, a linear--Gaussian SCM, an intervention budget, and a fixed observational sample. The environment hides $G$ and $\mathcal{M}$ and exposes only $D^{\mathrm{obs}}$, $d$, and $B$.

\subsection{World generation}

Each instance assumes causal sufficiency, linear--Gaussian noise (Eq.~\ref{eq:linear-gaussian}), and edge weights drawn uniformly from $[-2,-0.5]\cup[0.5,2]$. We sample a uniform topological order, select exactly $k$ forward edges, and randomly permute node indices so names do not leak topology. Instances are rejected if any relevant partial correlation falls below $\texttt{faithfulness\_eps}=0.1$. For each accepted instance we precompute the minimum single-node intervention set $\mathcal{I}^\star$ that orients every CPDAG-undirected edge; this defines the efficiency baseline.

\begin{figure}[t]
\centering
\includegraphics[width=0.72\linewidth]{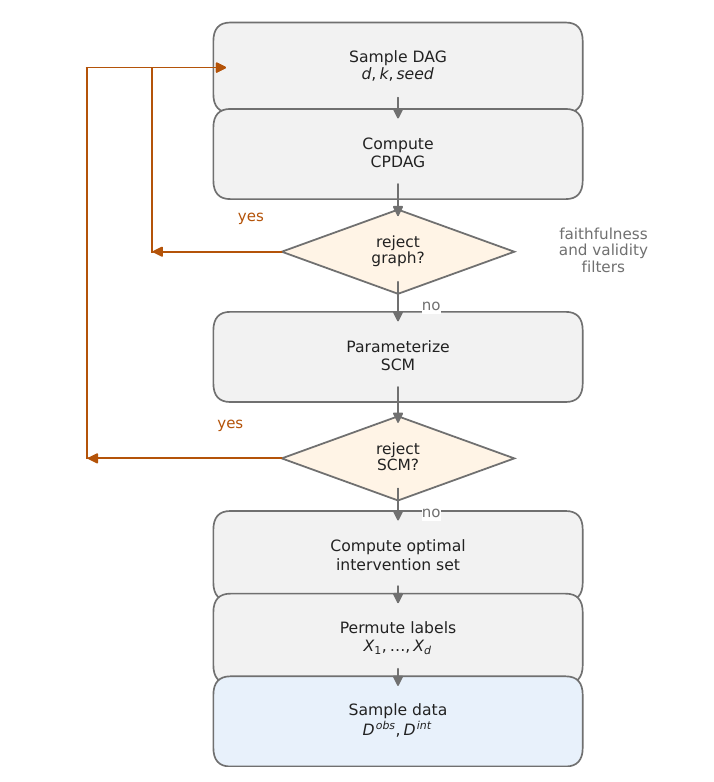}
\caption{ACDB instance generation. Each accepted instance is produced by sampling a graph, converting it to its CPDAG, rejecting invalid or near-unfaithful cases, parameterizing a linear--Gaussian SCM, computing the optimal intervention set, anonymizing labels, and sampling data. Rejection gates loop back to graph sampling.}
\label{fig:instance-generation}
\vspace{-0.6em}
\end{figure}

\subsection{Session API}

An ACDB session enforces a strict observe--intervene--submit protocol. \texttt{observe()} returns the pre-sampled $D^{\mathrm{obs}}$ once. \texttt{intervene(var, value)} draws $n_{\mathrm{int}}$ fresh samples under the hard intervention $\mathrm{do}(X_{\texttt{var}} = \texttt{value})$ and decrements the budget; requests beyond $B$ fail. \texttt{submit\_graph(submission)} seals the session. Submissions are mixed graphs of directed and undirected edges; undirected edges represent orientations the agent was unable or unwilling to commit to. Interventions are hard (the target variable is set exogenously; its parents are severed during sampling), and the scoring contract treats an unresolved edge as a distinct failure mode from a reversal or omission.

\begin{figure}[t]
\centering
\includegraphics[width=\linewidth]{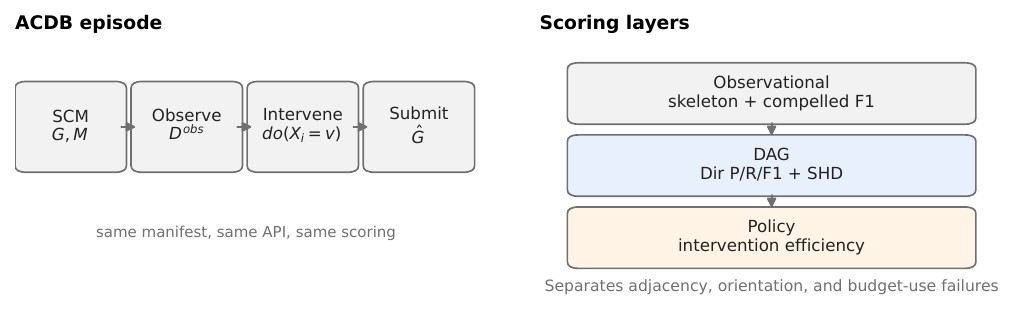}
\caption{ACDB fixes the episode protocol and separates adjacency, orientation, and budget-use scores.}
\label{fig:protocol-scoring}
\vspace{-0.6em}
\end{figure}

\subsection{Scoring contract}
\label{sec:scoring}

ACDB reports three complementary views of a submission; all are deterministic given $(G,\text{submission},B)$.

\paragraph{Observational layer: skeleton and compelled F1.} Against the true CPDAG of $G$, we compute standard precision/recall/F1 on the undirected skeleton ($\mathit{skeleton\_f1}$) and on the compelled (observationally orientable) edges ($\mathit{compelled\_f1}$). A reversed compelled edge counts as both a false positive on the reversed direction and a false negative on the true direction, so the agent is not rewarded for ``at least the adjacency was right'' when it flipped a compelled orientation.

\paragraph{DAG layer: directed F1 and SHD.} Against the true DAG $G$ itself, we compute set-intersection precision/recall/F1 on directed edges ($\mathit{directed\_f1}$) and a structural Hamming distance $\mathit{dag\_shd}$. SHD counts, over the union of true and submitted adjacencies, the following errors at one per edge: \emph{extra} (submission has an edge, truth does not), \emph{missing} (truth has an edge, submission has nothing), \emph{reversed} (both have a directed edge but in opposite directions), and \emph{unresolved} (truth has a directed edge, submission has an undirected edge on the same pair). Note that a reversal is one error, not two.

\paragraph{Efficiency layer.} Given the precomputed optimal intervention set $\mathcal{I}^\star$ of size $|\mathcal{I}^\star|$ and the number of interventions actually used, the efficiency score is
\begin{equation}
\eta \;=\; \frac{|\mathcal{I}^\star|}{\max(|\mathcal{I}_{\mathrm{used}}|, |\mathcal{I}^\star|)}.
\label{eq:efficiency}
\end{equation}
This is one whenever the agent used at most the optimal number of interventions (including zero-budget submission when $\mathcal{I}^\star = \varnothing$), and decreases toward zero as wasted interventions accumulate.

\paragraph{What the scoring contract excludes.} There is no MEC-aware partial credit in the DAG layer: the observational layer already expresses MEC-ceiling performance, and making the DAG layer MEC-aware would collapse the two.

\subsection{The difficulty ladder}

ACDB organizes instances into six levels, summarized in Table~\ref{tab:ladder}. Each level is a triple of difficulty axes: graph size and density, statistical power (sample counts and noise variance), and budget tightness (\texttt{budget\_slack} controls how many interventions above $|\mathcal{I}^\star|$ are allotted). The ladder is not a single monotone axis; levels 2 and 4 are designed to hold graph size fixed and isolate one axis at a time (statistical in L2, budget in L4), while L3 and L5 are structural stress levels.

\begin{table}[t]
\centering
\caption{ACDB ladder. Each level fixes $d$ (variables), $k$ (true edge count), $n_{\mathrm{obs}}$ / $n_{\mathrm{int}}$ (sample counts), $\sigma^2$ (noise variance), and \texttt{slack} (interventions above the precomputed optimal). The intervention budget is $B = |\mathcal{I}^\star| + \texttt{slack}$ and is therefore instance-specific.}
\label{tab:ladder}
\begin{tabular}{cllcccccc}
\toprule
Level & Role & $d$ & $k$ & $n_{\mathrm{obs}}$ & $n_{\mathrm{int}}$ & $\sigma^2$ & slack \\
\midrule
0 & tutorial       & 4 & 5 & 50 & 25 & 0.5 & 2 \\
1 & standard       & 5 & 6 & 25 & 15 & 1.0 & 1 \\
2 & statistical    & 5 & 6 & 15 & 10 & 1.5 & 1 \\
3 & structural     & 7 & 9 & 25 & 15 & 1.0 & 1 \\
4 & pressure       & 5 & 6 & 25 & 15 & 1.0 & 0 \\
5 & hard           & 7 & 9 & 15 & 10 & 1.5 & 0 \\
\bottomrule
\end{tabular}
\end{table}

Each level is evaluated on 8 seeds, preflight-validated so that instance assembly (DAG sample $\to$ faithfulness check $\to$ SCM parameterization $\to$ optimal intervention set) succeeds for all seeds in the manifest. The same manifest is used across all agents and baselines for a given ladder run, so comparisons are paired.

\subsection{Density leakage: a benchmark-design hazard}
\label{sec:density}

A structure-blind random baseline --- one that samples a random DAG using only the public variable count $d$, with no access to the true edge count $k$ --- provides a floor for interpreting agent performance. When graph density $\rho = k / \binom{d}{2}$ is high, this floor is also high: a randomly sampled DAG agrees with the truth on a non-trivial fraction of directed edges simply because both contain many edges. We call this effect \emph{density leakage}.

Empirically, on the ladder of Table~\ref{tab:ladder}, the structure-blind random baseline attains overall directed F1 of $0.236$ across $4800$ samples (Section~\ref{sec:results}). Measuring the same baseline at several post hoc densities, without giving the baseline access to $k$, we observe an approximately monotone heuristic:
\begin{equation}
\mathbb{E}[\text{Dir-F1}_{\text{random}}] \;\approx\; \frac{\rho}{1 + 2\rho}, \qquad \rho = \frac{k}{\binom{d}{2}}.
\label{eq:density-leakage}
\end{equation}
Equation~\eqref{eq:density-leakage} is a probe fit, not a derived bound. The density-probe ladder ($\rho \le 0.33$ on non-tutorial levels) reduces the measured random floor from $23.6\%$ to $16.9\%$ overall. We keep the original ladder for v0 because the LLM and PC runs were collected there, and use the probe as a concrete target for v1 calibration.

\begin{figure}[t]
\centering
\includegraphics[width=0.95\linewidth]{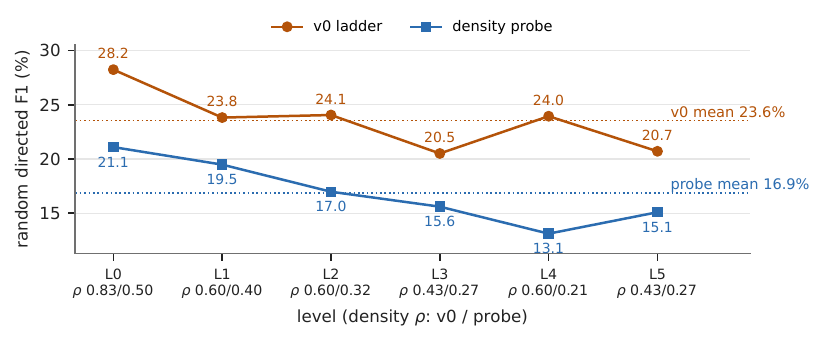}
\caption{Density leakage in the structure-blind random baseline. Points are per-level directed F1; tick labels show graph density $\rho$ for the v0 ladder and density probe. Lowering density reduces the overall random floor from $23.6\%$ to $16.9\%$.}
\label{fig:density-leakage}
\vspace{-0.6em}
\end{figure}

\section{Agents and Baselines}
\label{sec:agents}

All agents interact through the same session API (\S\ref{sec:benchmark}), see the same manifest of $8$ preflighted seeds per level, and are scored by the same three-tier contract. Agents differ only in the policy inside the session.

\paragraph{PC and PC + greedy active.} The observational baseline runs PC from the \texttt{causal-learn} library on $D^{\mathrm{obs}}$ with Fisher-$z$ independence tests at $\alpha = 0.05$ and submits the resulting CPDAG. The active baseline extends this procedure: after PC, it repeatedly selects the node with the highest undirected-edge degree, intervenes at $\mathrm{do}(X_i = \bar{X}_i + 3.0)$ where $\bar{X}_i$ is the observational mean, and orients each unresolved neighbor by comparing the post-intervention mean against the observational mean (shift threshold $0.5$ in absolute value). Cycles are rejected. The greedy heuristic is simple and not optimal, but it provides a clear reference policy for the decisions an LLM policy must make. An \emph{oracle} baseline submits the true DAG as an upper bound.

\paragraph{Structure-blind random baseline.} The random-uniform baseline quantifies how much directed-F1 signal is achievable without using data. Using only the public variable count $d$ (not $D^{\mathrm{obs}}$, not $k$, no interventions), it samples a random topological order, enumerates the $M = d(d-1)/2$ forward edges, draws an edge count $m$ uniformly from $\{0, \dots, M\}$, and submits $m$ random forward edges. We run $100$ such samples per instance. Results near this floor should not be interpreted as evidence of reliable causal discovery (\S\ref{sec:density-results}).

\paragraph{LLM agents.} Two backbones (GPT-5.4 and Claude Sonnet 4.6), four configurations each: raw (\texttt{llm\_raw}) vs stats (\texttt{llm\_stats}), on both observational and active panels. Temperature is $0$. Each turn the model receives the task prompt, a static session prompt (variable names, observational matrix, initial budget), and the full tool-call history, and emits exactly one tool call from $\{$\texttt{intervene}, \texttt{correlation}, \texttt{partial\_correlation}, \texttt{independence\_test}, \texttt{submit\_graph}$\}$. Parallel tool calls are disabled. Episodes end at \texttt{submit\_graph} or at the step limit ($16$ for raw, $32$ for stats). The raw variant must reason over the raw data matrix; the stats variant additionally has correlation, partial-correlation, and Fisher-$z$ tools. The raw/stats comparison tests whether handing the model statistical primitives closes the gap to PC; the observational/active split measures whether the model can use interventions beyond the CPDAG ceiling.

\paragraph{Correlation-only ablation probe.} A restricted-prompt probe (\texttt{llm\_corr\_obs}) gives the model only the correlation matrix (rounded to three decimal places) and sample standard deviations, with no raw data, no tools, no interventions, and a system prompt that warns that correlation does not imply causation and asks the model to withhold orientation when uncertain. The model must respond with a single \texttt{submit\_graph} call. We run this probe on levels $\{1, 3, 5\}$ with $4$ seeds per level and both backbones. It is a diagnostic ablation, not a primary comparison (\S\ref{sec:corr-probe}).

\section{Results}
\label{sec:results}

We evaluate three families of agents on the ladder of Table~\ref{tab:ladder}: the PC algorithm (classical constraint-based baseline), LLM policies (GPT-5.4 and Claude Sonnet 4.6, each in raw-data and statistical-tools variants), and two reference probes (a structure-blind random DAG baseline and a correlation-only LLM ablation). The run comprised $336$ planned jobs per model across the full ladder; GPT-5.4 completed all $336$ successfully, Sonnet 4.6 completed $335$ (one \texttt{llm\_stats} active row failed). All ladder-evaluated agents share the same manifest of $8$ seeds per level, so every pairwise comparison is on identical instances.

\subsection{Headline results}

Table~\ref{tab:ladder-overall} reports weighted averages across L0--L5 with directed precision and recall broken out alongside F1. Paired sign-flip tests on deduplicated same-seed rows support the main active comparisons: PC+greedy beats the best LLM on directed F1 by $11.0$ points ($p=0.025$) and SHD by $2.46$ edges ($p<10^{-4}$), while Sonnet raw beats GPT raw in the active panel by $8.8$ points ($p=0.028$). Four findings summarize the run.

\begin{figure}[t]
\centering
\includegraphics[width=0.95\linewidth]{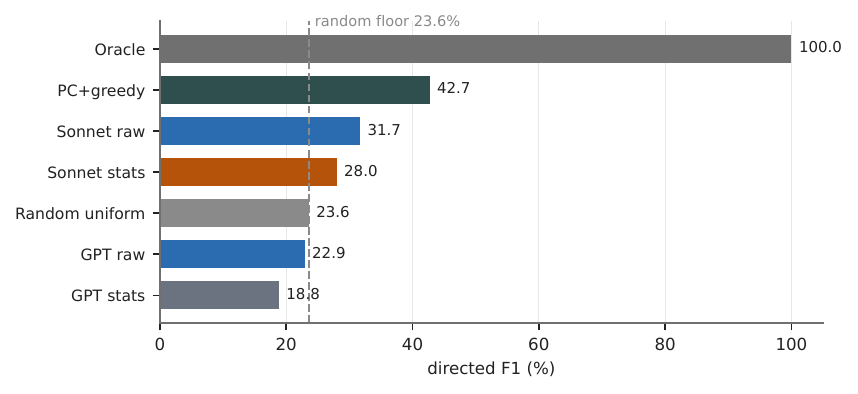}
\caption{Headline DAG-recovery performance on the v0 active ladder. PC+greedy is the strongest non-oracle method, Sonnet raw is the strongest LLM policy, and several LLM settings sit close to the structure-blind random floor.}
\label{fig:headline-directed-f1}
\vspace{-0.6em}
\end{figure}

\begin{table}[t]
\centering
\caption{Headline results as weighted averages across ladder levels L0--L5.
Skel~F1 is against the true CPDAG; Dir~P, Dir~R, Dir~F1 are against the true
DAG; SHD is edge edit distance (reversal=1); $\eta$ is intervention efficiency
against the precomputed optimal set. All values in \%; SHD is a mean edge count.
Oracle submits the true DAG; PC + greedy is the strongest non-oracle active
method. Directed precision and recall are reported separately because LLM
and PC errors lie on different sides of the P/R trade-off (\S\ref{sec:results}).}
\label{tab:ladder-overall}
\small
\begin{tabular}{llcccccc}
\toprule
Method & Model & Skel~F1 & Dir~P & Dir~R & Dir~F1 & SHD & $\eta$ \\
\midrule
\multicolumn{8}{l}{\textit{Observational panel (no interventions)}} \\
\quad PC                 & --                 & 62.1 & 82.6 & 10.0 & 13.6 & 6.25 & -- \\
\quad LLM (raw)          & GPT-5.4            & 65.7 & 74.1 & 19.2 & 22.6 & 9.40 & -- \\
\quad LLM (stats)        & GPT-5.4            & 35.9 & 68.4 &  9.2 & 13.3 & 6.75 & -- \\
\quad LLM (raw)          & Sonnet 4.6         & 57.7 & 38.9 & 34.5 & \textbf{36.1} & 7.46 & -- \\
\quad LLM (stats)        & Sonnet 4.6         & 60.7 & 76.9 & 12.5 & 15.0 & 9.92 & -- \\
\midrule
\multicolumn{8}{l}{\textit{Active panel (hard-intervention budget)}} \\
\quad Oracle             & --                 & 100.0 & 100.0 & 100.0 & 100.0 & 0.00 & 100.0 \\
\quad PC + greedy        & --                 & 62.1 & 66.9 & 32.8 & \textbf{42.7} & \textbf{4.79} & 89.6 \\
\quad LLM (raw)          & GPT-5.4            & 58.8 & 36.3 & 19.3 & 22.9 & 9.27 & 92.7 \\
\quad LLM (stats)        & GPT-5.4            & 40.0 & 48.0 & 13.3 & 18.8 & 6.58 & 100.0 \\
\quad LLM (raw)          & Sonnet 4.6         & 56.3 & 35.0 & 29.6 & 31.7 & 7.25 & 97.9 \\
\quad LLM (stats)        & Sonnet 4.6         & 62.7 & 46.0 & 22.7 & 28.0 & 8.06 & 100.0 \\
\midrule
\multicolumn{8}{l}{\textit{Reference points}} \\
\quad Random uniform DAG & --                 & 46.6 & 35.0 & 25.0 & 23.6 & 8.18 & -- \\
\quad Correlation-only probe\textsuperscript{\dag} & GPT-5.4, Sonnet 4.6 & $\sim$70 & -- & -- & 0.0 & -- & -- \\
\bottomrule
\end{tabular}

\vspace{2pt}
\begin{flushleft}
\footnotesize
\textsuperscript{\dag}Prompt-restricted ablation: single-call graph submission
from the correlation matrix only, with no interventions or tool calls.
Skeleton F1 $\approx 70\%$; directed F1 $= 0$ under the prompt restriction,
because the models withheld orientation when uncertain.
Not a primary comparison. See Section~\ref{sec:corr-probe}.
\end{flushleft}
\end{table}

\textbf{Finding 1: PC with active orientation is strongest on the primary metrics.} PC + greedy achieved directed F1 of $42.7\%$ and SHD of $4.79$, outperforming every LLM agent. The best LLM (Claude Sonnet 4.6 active raw) reached directed F1 of $31.7\%$ and SHD of $7.25$ under the same budget. The SHD gap is the more robust of the two because it is less sensitive to aggressive guessing.

\textbf{Finding 2: the P/R decomposition reveals qualitatively different errors.} On directed edges in the active panel, PC + greedy had precision $66.9\%$ and recall $32.8\%$: it commits to few edges but is usually correct when it does. Sonnet 4.6 raw was more balanced ($35.0/29.6$), while GPT-5.4 raw submitted aggressively with low precision ($36.3/19.3$). This is the main qualitative finding: \emph{PC under-commits, LLMs over-commit, and F1 alone collapses both errors into a single number.} Reporting only F1 would understate how different the two failure modes are.

\textbf{Finding 3: Sonnet 4.6 is stronger than GPT-5.4 in every overall LLM setting.} Directed F1 improved for Sonnet 4.6 across raw and stats, observational and active. The gap was largest on raw observational ($36.1\%$ vs $22.6\%$, $p=0.0028$) and raw active ($31.7\%$ vs $22.9\%$, $p=0.028$). Under these prompts and seeds, Sonnet 4.6's observational panel matched or exceeded every GPT-5.4 setting, including active, on directed F1.

\textbf{Finding 4: statistical-tool access shifted behavior toward abstention.} Agents given correlation, partial correlation, and Fisher-$z$ tools did not outperform raw-data agents on directed F1: Sonnet raw beat Sonnet stats ($31.7\%$ vs $28.0\%$), and GPT raw beat GPT stats ($22.9\%$ vs $18.8\%$). Stats agents ran longer trajectories, intervened less, and submitted fewer directed edges; their high efficiency therefore reflects abstention rather than better active discovery.

\subsection{Per-level breakdown}

Table~\ref{tab:ladder-perlevel} shows directed F1 per level for both panels. Three patterns are visible.

First, PC + greedy is strongest overall but not uniformly across levels. Sonnet raw exceeds it on L2 ($36.9\%$ vs $30.7\%$) and L5 ($31.9\%$ vs $15.2\%$), the two low-sample levels where PC's finite-sample errors are largest.

Second, Sonnet 4.6 is the stronger LLM in the overall aggregates, but not every individual cell. GPT-5.4 occasionally lands near the random floor, especially in the observational panel at L2 and L5.

Third, the \emph{observational} panel outscores the \emph{active} panel for Sonnet 4.6 raw at L2 ($39.3\%$ vs $36.9\%$) and L5 ($40.1\%$ vs $31.9\%$). These paired gaps are not significant at eight seeds ($p=0.50$, $p=0.20$), but they are diagnostically useful: intervention data did not reliably improve the policy on the hardest levels.

\begin{table}[t]
\centering
\scriptsize
\setlength{\tabcolsep}{3.2pt}
\renewcommand{\arraystretch}{0.92}
\caption{Directed F1 per ladder level, by panel. Weighted averages in the rightmost column match Table~\ref{tab:ladder-overall}. Values are means over $8$ preflighted seeds. The random row is structure-blind and uses only $d$. All values in \%.}
\label{tab:ladder-perlevel}
\begin{tabular}{llcccccc|c}
\toprule
Method & Model & L0 & L1 & L2 & L3 & L4 & L5 & Overall \\
\midrule
\multicolumn{9}{l}{\textit{Observational panel}} \\
PC              & --                & 13.9 & 23.8 & 3.1  & 17.2 & 20.0 & 3.8  & 13.6 \\
LLM (raw)       & GPT-5.4           & 39.8 & 27.7 & 7.5  & 18.8 & 24.4 & 17.7 & 22.6 \\
LLM (stats)     & GPT-5.4           & 12.3 & 13.4 & 9.4  & 11.4 & 21.9 & 11.8 & 13.3 \\
LLM (raw)       & Sonnet 4.6        & 26.9 & 34.9 & 39.3 & 33.8 & 41.7 & 40.1 & \textbf{36.1} \\
LLM (stats)     & Sonnet 4.6        & 24.7 & 14.0 & 6.2  & 10.9 & 22.4 & 12.0 & 15.0 \\
\midrule
\multicolumn{9}{l}{\textit{Active panel}} \\
Oracle          & --                & 100.0 & 100.0 & 100.0 & 100.0 & 100.0 & 100.0 & 100.0 \\
PC + greedy     & --                & \textbf{65.1} & \textbf{49.7} & 30.7 & \textbf{48.9} & \textbf{46.5} & 15.2 & \textbf{42.7} \\
LLM (raw)       & GPT-5.4           & 28.5 & 24.1 & 27.6 & 20.6 & 15.0 & 21.6 & 22.9 \\
LLM (stats)     & GPT-5.4           & 31.2 & 13.4 & 12.2 & 18.3 & 21.6 & 16.4 & 18.8 \\
LLM (raw)       & Sonnet 4.6        & 29.4 & 28.4 & \textbf{36.9} & 29.7 & 33.6 & \textbf{31.9} & 31.7 \\
LLM (stats)     & Sonnet 4.6        & 26.4 & 33.2 & 29.7 & 28.7 & 26.5 & 23.5 & 28.0 \\
\midrule
Random uniform  & --                & 28.2 & 23.8 & 24.1 & 20.5 & 24.0 & 20.7 & 23.6 \\
\bottomrule
\end{tabular}
\renewcommand{\arraystretch}{1.0}
\end{table}

\subsection{Density leakage: the random floor}
\label{sec:density-results}

The structure-blind random baseline attained overall directed F1 of $23.6\%$ on the ladder, with per-level values from $28.2\%$ at L0 down to about $21\%$ at structural levels. Several GPT-5.4 settings land near this floor (raw active $22.9\%$, raw observational $22.6\%$, stats active $18.8\%$), so their directed-F1 margins should be treated cautiously. A separate density probe, using the same random policy without giving it $k$, reduces the random floor to $16.9\%$ overall. This is the empirical reason v1 should recalibrate density before drawing stronger conclusions about LLM capability.

\subsection{Correlation-only ablation probe}
\label{sec:corr-probe}

As a diagnostic, we ran \texttt{llm\_corr\_obs} on L1/L3/L5 with $4$ seeds per level and both backbones. The model saw only standard deviations and a rounded correlation matrix, no raw data, tools, or interventions. Skeleton F1 was non-trivial ($67$--$74\%$ by level/model), showing that compact summaries carry adjacency signal in this SCM family. Directed F1 was $0$ because the prompt explicitly asked the models to withhold unsupported orientations and they submitted undirected edges. This is an incomplete v0 ablation, not a primary DAG-recovery comparison.

\subsection{Efficiency: budgets were rarely the binding constraint}

PC + greedy achieved efficiency $\eta=89.6\%$; it sometimes used redundant interventions. LLM agents achieved $\eta\in[92.7,100.0]\%$, with both \texttt{llm\_stats} settings at exactly $100.0\%$. This is not skilled budget management: the stats agents rarely intervened, preferring statistical tool calls and conservative submissions. High efficiency without matching precision/recall gains is evidence of abstention.

\subsection{Summary}

PC + greedy is strongest overall because it combines high directed precision with a narrow commit pattern. LLMs commit more aggressively and with lower precision. Sonnet 4.6 is the stronger overall LLM under these prompts, statistical-tool access does not improve directed F1, and current LLM policies do not reliably convert intervention budget into orientation accuracy.

\section{Discussion}
\label{sec:discussion}

We interpret the main findings and identify the calibration work needed for the next benchmark version.

\subsection{What the results say about current LLMs}

\paragraph{Classical methods still win on clean formal tasks.} PC + greedy achieved the highest directed F1 and the lowest SHD on the ladder. Under the v0 assumptions, this is consistent with theory: PC is asymptotically correct and the greedy orientation heuristic is strong enough for small graphs. Current LLM agents are not drop-in replacements for classical structure-learning algorithms in settings where the SCM lies in their identifiability regime.

\paragraph{The precision--recall asymmetry is the main behavioral signal.} PC under-commits (precision $66.9$, recall $32.8$); LLMs over-commit (precision $35$--$46$, recall $20$--$30$). Neither failure mode is uniformly preferable: under-committers leave useful edges unresolved, while over-committers add wrong orientations. A benchmark that reports only F1 collapses this distinction. ACDB's scoring contract keeps precision and recall visible.

\paragraph{Interventions sometimes hurt.} At L2 and L5, Sonnet 4.6's observational directed F1 exceeded its active directed F1 on identical seeds ($39.3\%$ vs $36.9\%$; $40.1\%$ vs $31.9\%$). Interventions are more informative than observations, so this is a policy diagnostic: current prompts do not reliably convert intervention samples into better orientation decisions. Edge-level revision analysis is a v1 target.

\paragraph{Statistical tools did not improve active discovery.} Giving LLMs correlation, partial-correlation, and independence-test tools did not raise directed F1 above raw data. The stats variants ran longer trajectories, intervened less, and submitted fewer directed edges; their $100\%$ efficiency reflects abstention, not skill. Tool access alone is insufficient if the model does not know when to stop testing and commit.

\subsection{What the results say about the benchmark}

\paragraph{Density leakage is an open calibration issue.} The random baseline's $23.6\%$ directed F1 on the current ladder is high enough that several GPT-5.4 settings sit near it. The density probe reduces the random floor to $16.9\%$ overall, turning the ladder issue into a concrete v1 calibration target rather than a vague caveat. The v0 results remain useful because all agents share paired seeds on the same ladder.

\paragraph{The scoring contract separates distinct failure modes.} The decomposition into skeleton, compelled-edge, DAG, SHD, and efficiency scores separated the CPDAG-ceiling question from the orientation-under-intervention question from the policy-quality question. The correlation-only probe would not be legible in an F1-only benchmark: its zero directed F1 looks like a failure, but the prompt-restricted reading --- the model withheld orientation when uncertain --- is only available because skeleton F1 is reported alongside.

\subsection{v1 and future variants}
\label{sec:future-scm-ladder}

The next benchmark version should keep the same SCM foundation but fix the density ladder, run the full LLM-correlation summary condition, increase seeds, and inspect intervention-driven revision behavior. Beyond that, the same generator/API can support causal inference with a known DAG, counterfactual queries, decision-making under causal uncertainty, transfer across SCM families, and RL for experimental design. Those are extensions of the scaffold, not claims made by this v0 run.

\subsection{Limitations}

Four limitations are central to interpreting the results. (1) v0 assumes causal sufficiency and linear--Gaussian SCMs. (2) The ladder has density-leakage artifacts at the dense end. (3) Eight seeds per level is adequate for coarse trends but under-powered for fine-grained ranking. (4) We tested two LLM backbones and one prompt design each, so LLM capability claims are conditional on those choices.

\section{Related Work}
\label{sec:related}

\paragraph{Classical causal discovery.} ACDB builds on the PC algorithm \citep{spirtes2000causation}, Markov-equivalence theory \citep{verma1990equivalence,andersson1997characterization}, score-based GES \citep{chickering2002ges}, and intervention-design work on orienting equivalence classes \citep{eberhardt2005interventions,hauser2012gies}. These methods define the CPDAG ceiling and the classical baselines against which LLM agents should be evaluated.

\paragraph{LLMs for causality.} Prior LLM work ranges from optimistic pairwise causal-direction results \citep{kiciman2023llmcausal} to critiques that such behavior is pattern matching rather than causal reasoning \citep{zecevic2023causalparrots}. CLadder evaluates language-model causal QA across Pearl's ladder \citep{jin2023cladder}, while \citet{long2023llmgraphs} test text-only graph reconstruction. ACDB differs by requiring the model to interact with data and interventions before committing to a scored graph.

\paragraph{Agent benchmarks.} SWE-bench, AgentBench, WebArena, OSWorld, and GAIA move evaluation toward interactive agents \citep{jimenez2024swebench,liu2024agentbench,zhou2024webarena,xie2024osworld,mialon2023gaia}. ACDB shares the paired-seed, deterministic-scoring ethos but narrows the task to scientific reasoning under a formal SCM, where the hidden state and failure modes are inspectable by construction.

\section{Conclusion}
\label{sec:conclusion}

We asked whether an LLM agent, given budgeted hard interventions, can recover a causal DAG better than a structure-blind random baseline. On the v0 ladder, PC + greedy is strongest overall ($42.7\%$ directed F1, SHD $4.79$); the best LLM reaches $31.7\%$ directed F1 with SHD $7.25$; and the random baseline reaches $23.6\%$, high enough that weak LLM margins require caution. The precision--recall decomposition shows qualitatively different errors: PC under-commits with high precision, LLMs over-commit with lower precision, and stats-tool variants often abstain from intervention. ACDB is therefore a scaffold for tracking active causal discovery as models and benchmark calibration improve. The next version should lower the random floor, run the full correlation-summary condition, and inspect intervention-driven revisions.

\bibliographystyle{plainnat}
\bibliography{references}

\end{document}